\documentclass[11pt]{article}
\usepackage{coling2018}
\usepackage{times}
\usepackage{url}
\usepackage{latexsym}

\title{ Improving Neural Text Simplification Model with Simplified Corpora}

\author{Jipeng Qiang \\
  Department of Computer Science, Yangzhou University/ Yangzhou, Jiangsu, China  \\
  {\tt jpqiang@yzu.edu.cn} \\}

\date{}

\begin{document}
\maketitle
\begin{abstract}
   
  Text simplification (TS) can be viewed as monolingual translation task, translating between text variations within a single language. Recent neural TS models draw on insights from neural machine translation to learn lexical simplification and content reduction using encoder-decoder model. But different from neural machine translation, we cannot obtain enough ordinary and simplified sentence pairs for TS, which are expensive and time-consuming to build. Target-side simplified sentences plays an important role in boosting fluency for statistical TS, and we investigate the use of simplified sentences to train, with no changes to the network architecture. We propose to pair simple training sentence with a synthetic ordinary sentence via back-translation, and treating this synthetic data as additional training data. We train encoder-decoder model using synthetic sentence pairs and original sentence pairs, which can obtain substantial improvements on the available WikiLarge data and WikiSmall data compared with the state-of-the-art methods. 

\end{abstract}

\section{Introduction}

Text simplification aims to reduce the lexical and structural complexity of a text, while still retaining the semantic meaning, which can help children, non-native speakers, and people with cognitive disabilities, to understand text better. 
One of the methods of automatic text simplification can be generally divided into three categories: lexical simplification (LS) \cite{biran2011putting,paetzold2016unsupervised}, rule-based \cite{vstajner2017sentence}, and machine translation (MT) \cite{zhu2010monolingual,wang2016experimental}. LS is mainly used to simplify text by substituting infrequent and difficult words with frequent and easier words. However, there are several challenges for the LS approach: a great number of transformation rules are required for reasonable coverage and should be applied based on the specific context; third, the syntax and semantic meaning of the sentence is hard to retain. Rule-based approaches use hand-crafted rules for lexical and syntactic simplification, for example, substituting difficult words in a predefined vocabulary. However, such approaches need a lot of human-involvement to manually define these rules, and it is impossible to give all possible simplification rules. MT-based approach has attracted great attention in the last several years, which addresses text simplification as a monolingual machine translation problem translating from 'ordinary' and 'simplified' sentences. 

In recent years, neural Machine Translation (NMT) is a newly-proposed deep learning approach and achieves very impressive results \cite{bahdanau2014neural,rush2015neural,sutskever2014sequence}. Unlike the traditional phrased-based machine translation system which operates on small components separately, NMT system is being trained end-to-end, without the need to have external decoders, language models or phrase tables. Therefore, the existing architectures in NMT are used for text simplification \cite{nisioi2017exploring,wang2016experimental}. However, most recent work using NMT is limited to the training data that are scarce and expensive to build. Language models trained on simplified corpora have played a central role in statistical text simplification \cite{glavavs2015simplifying,coster2011simple}. One main reason is the amount of available simplified corpora typically far exceeds the amount of parallel data. The performance of models can be typically improved when trained on more data. Therefore, we expect simplified corpora to be especially helpful for NMT models.

In contrast to previous work, which uses the existing NMT models, we explore strategy to include simplified training corpora in the training process without changing the neural network architecture. We first propose to pair simplified training sentences with synthetic ordinary sentences during training, and treat this synthetic data as additional training data. We obtain synthetic ordinary sentences through back-translation, i.e. an automatic translation of the simplified  sentence into the ordinary sentence \cite{sennrich2015improving}. Then, we mix the synthetic data into the original (simplified-ordinary) data to train NMT model. Experimental results on two publicly available datasets show that we can improve the text simplification quality of NMT models by mixing simplified sentences into the training set over NMT model only using the original training data.

\section{Related Work}

Automatic TS is a complicated natural language processing (NLP) task, which consists of lexical and syntactic simplification levels \cite{Kauchak2013Improving}. It has attracted much attention recently as it could make texts more accessible to wider
audiences, and used as a pre-processing step, improve performances of various NLP tasks and systems \cite{qiang2018snapshot,qiang2018sttm,qiang2018short}.
Usually, hand-crafted, supervised, and unsupervised methods based on resources like English Wikipedia and Simple English Wikipedia (EW-SEW) \cite{coster2011simple} are utilized for extracting simplification rules. It is very easy to mix up the automatic TS task and the automatic summarization task \cite{zhu2010monolingual,hwang2015aligning,rush2015neural}. TS is different from text summarization as the focus of text summarization is to reduce the length and redundant content.

At the lexical level, lexical simplification systems often substitute difficult words using more common words, which only require a large corpus of regular text to obtain
word embeddings to get words similar to the complex word \cite{paetzold2016unsupervised,glavavs2015simplifying}. Biran et al. \cite{biran2011putting} adopted an unsupervised method for learning pairs of complex and simpler synonyms from a corpus consisting of Wikipedia and Simple Wikipedia. At the sentence level, a sentence simplification model was proposed by tree transformation based on statistical machine translation (SMT) \cite{zhu2010monolingual}. Woodsend and Lapata \cite{woodsend2011learning} presented a data-driven model based on a quasi-synchronous grammar, a formalism that can naturally capture structural mismatches and complex rewrite operations. Wubben et al. \cite{wubben2012} proposed a phrase-based machine translation (PBMT) model that is trained on ordinary-simplified sentence pairs. Xu et al. \cite{xu2016optimizing} proposed a syntax-based machine translation model using simplification-specific objective functions and features to encourage simpler output. 

Compared with SMT, neural machine translation (NMT) has shown to produce state-of-the-art results \cite{bahdanau2014neural,sutskever2014sequence}. The central approach of NMT is an encoder-decoder architecture implemented by recurrent neural networks, which can represent the input sequence as a vector, and then decode
that vector into an output sequence. Therefore, NMT models were used for text simplification task, and achieved good results \cite{nisioi2017exploring,wang2016experimental,zhang2017sentence}. The main limitation of the aforementioned NMT models for text simplification depended on the parallel ordinary-simplified sentence pairs. Because ordinary-simplified sentence pairs are expensive and time-consuming to build, the available largest data is EW-SEW that only have 296,402 sentence pairs. The dataset is insufficiency for NMT model if we want to NMT model can obtain the best parameters. Considering simplified data plays an important role in boosting fluency for phrase-based text simplification, and we investigate the use of simplified data for text simplification. We are the first to show that we can effectively adapt neural translation models for text simplifiation with simplified corpora.

\section{NMT Training with Simplified Corpora}

\subsection{Simplified Corpora}

We collected a simplified dataset from Simple English Wikipedia that are freely available\footnote{simple.wikipedia.org}, which has been previously used for many text simplification methods \cite{biran2011putting,coster2011simple,zhu2010monolingual}. The simple English Wikipedia is pretty easy to understand than normal English Wikipedia. We downloaded all articles from Simple English Wikipedia. For these articles, we removed stubs, navigation pages and any article that consisted of a single sentence. We then split them into sentences with the Stanford CorNLP \cite{Manning2014}, and deleted these sentences whose number of words are smaller than 10 or large than 40. After removing repeated sentences, we chose 600K sentences as the simplified data with 11.6M words, and the size of vocabulary is 82K.

\subsection{Text Simplification using Neural Machine Translation}

Our work is built on attention-based NMT \cite{bahdanau2014neural} as an encoder-decoder network with recurrent neural networks (RNN), which simultaneously conducts dynamic alignment and generation of the target simplified sentence. 

The encoder uses a bidirectional RNN that consists of forward and backward RNN. Given a source sentence $X=(x_1,x_2,...,x_l)$, the forward RNN and backward RNN calculate forward hidden states $(\overrightarrow{h_1},...,\overrightarrow{h_l})$ and backward hidden states $(\overleftarrow{h_1},...,\overleftarrow{h_l})$, respectively. The annotation vector $h_j$ is obtained by concatenating $\overrightarrow{h_j}$ and $\overrightarrow{h_j}$. 

The decoder is a RNN that predicts a target simplificated sentence with Gated Recurrent Unit (GRU) \cite{cho2014learning}. Given the previously generated target (simplified) sentence $Y=(y_1,y_2,...,y_{t-1})$, the probability of next target word $y_t$ is 

\begin{equation}
P(y_t|X) = softmax(g(e_{y_{t-1},s_t,c_t}))
\end{equation}
where $g(.)$ is a non-linear function, $e_{y_{t-1}}$ is the embedding of $y_{t-1}$, and $s_t$ is a decoding state for time step $t$.

State $s_t$ is calculated by 

\begin{equation}
s_t = f(s_{t-1},e_{y_{t-1}},c_t)
\end{equation}
where $f(.)$ is the activation function GRU. 

The $c_t$ is the context vector computed as a weighted annotation $h_j$, computed by 
\begin{equation}
c_t = \sum_j^{l}{a_{tj}\cdot h_j}
\end{equation}
where the weight $a_{tj}$ is computed by 
\begin{equation}
a_{tj} = \frac{exp(e_{tj})}{\sum_{i=1}^{l}{exp(e_{ti})}}
\end{equation}
\begin{equation}
e_{tj} = v_a^T tanh(W_as_{t-1}+U_ah_j)
\end{equation}
where $v_a$, $W_a$ and $U_a$ are weight matrices. The training objective is to maximize the likelihood of the training data. Beam search is employed for decoding.

\subsection{Synthetic Simplified Sentences}

We train an auxiliary system using NMT model from the simplified sentence to the ordinary sentence, which is first trained on the available parallel data. For leveraging simplified sentences to improve the quality of NMT model for text simplification, we propose to adapt the back-translation approach proposed by Sennrich et al.\cite{sennrich2015improving} to our scenario. More concretely, Given one sentence in simplified sentences, we use the simplified-ordinary system in translate mode with greedy decoding to translate it to the ordinary sentences, which is denoted as back-translation. This way, we obtain a synthetic parallel simplified-ordinary sentences. Both the synthetic sentences and the available parallel data are used as training data for the original NMT system. 

\section{Evaluation}

We evaluate the performance of text simplification using neural machine translation on available parallel sentences and additional simplified sentences.

\textbf{Dataset}. We use two simplification datasets (WikiSmall and WikiLarge). WikiSmall consists of ordinary and simplified sentences from the ordinary and simple English Wikipedias, which has been used as benchmark for evaluating text simplification \cite{woodsend2011learning,wubben2012,nisioi2017exploring}. The training set has 89,042 sentence pairs, and the test set has 100 pairs. WikiLarge is also from Wikipedia corpus whose training set contains 296,402 sentence pairs\cite{xu2016optimizing,zhang2017sentence}. WikiLarge includes 8 (reference) simplifications for 2,359 sentences split into 2,000 for development and 359 for testing.

\textbf{Metrics}. Three metrics in text simplification are chosen in this paper. BLEU \cite{bahdanau2014neural} is one traditional machine translation metric to assess the degree to which translated simplifications differed from reference simplifications. FKGL measures the readability of the output \cite{kincaid1975}. A small FKGL represents simpler output. SARI is a recent text-simplification metric by comparing the output against the source and reference simplifications \cite{zhang2017sentence}. 

We evaluate the output of all systems using human evaluation. The metric is denoted as Simplicity \cite{nisioi2017exploring}. The three non-native fluent English speakers are shown reference sentences and output sentences. They are asked whether the output sentence is much simpler (+2), somewhat simpler (+1), equally (0), somewhat more difficult (-1), and much more difficult (-2) than the reference sentence. 

\textbf{Methods}. We use OpenNMT \cite{opennmt} as the implementation of the NMT system for all experiments \cite{bahdanau2014neural}. We generally follow the default settings and training procedure described by Klein et al.(2017). We replace out-of-vocabulary words with a special UNK symbol. At prediction time, we replace UNK words with the highest probability score from the attention layer. OpenNMT system used on parallel data is the baseline system. To obtain a synthetic parallel training set, we back-translate a random sample of 100K sentences from the collected simplified corpora. OpenNMT used on parallel data and synthetic data is our model. The benchmarks are run on a Intel(R) Core(TM) i7-5930K CPU@3.50GHz, 32GB Mem, trained on 1 GPU GeForce GTX 1080 (Pascal) with CUDA v. 8.0.

We choose three statistical text simplification systems. PBMT-R is a phrase-based method with a reranking post-processing step \cite{wubben2012}. Hybrid performs sentence splitting and deletion operations based on discourse representation structures, and then simplifies sentences with PBMT-R \cite{Narayan2014}. SBMT-SARI \cite{xu2016optimizing} is syntax-based translation model using PPDB paraphrase database \cite{Ganitkevitch2013} and modifies tuning function (using SARI). We choose two neural text simplification systems. NMT is a basic attention-based encoder-decoder model which uses OpenNMT framework to train with two LSTM layers, hidden states of size 500 and 500 hidden units, SGD optimizer, and a dropout rate of 0.3 \cite{nisioi2017exploring}. Dress is an encoder-decoder model coupled with a deep reinforcement learning framework, and the parameters are chosen according to the original paper \cite{zhang2017sentence}. For the experiments with synthetic parallel data, we back-translate a random sample of 60 000 sentences from the collected simplified sentences into ordinary sentences. Our model is trained on synthetic data and the available parallel data, denoted as NMT+synthetic.

\textbf{Results}. Table 1 shows the results of all models on WikiLarge dataset. We can see that our method (NMT+synthetic) can obtain higher BLEU, lower FKGL and high SARI compared with other models, except Dress on FKGL and SBMT-SARI on SARI. It verified that including synthetic data during training is very effective, and yields an improvement over our baseline NMF by 2.11 BLEU, 1.7 FKGL and 1.07 SARI. We also substantially outperform Dress, who previously reported SOTA result. The results of our human evaluation using Simplicity are also presented in Table 1. NMT on synthetic data is significantly better than PBMT-R, Dress, and SBMT-SARI on Simplicity. It indicates that our method with simplified data is effective at creating simpler output.

Results on WikiSmall dataset are shown in Table 2. We see substantial improvements (6.37 BLEU) than NMT from adding simplified training data with synthetic ordinary sentences. Compared with statistical machine translation models (PBMT-R, Hybrid, SBMT-SARI), our method (NMT+synthetic) still have better results, but slightly worse FKGL and SARI. Similar to the results in WikiLarge, the results of our human evaluation using Simplicity outperforms the other models. In conclusion, Our method produces better results comparing with the baselines, which demonstrates the effectiveness of adding simplified training data.

\begin{table}
  \centering
  \caption{The evaluation results on WikiLarge. The best results are highlighted in bold on each metric.}
  \begin{tabular}{|c|c|c|c|c|} \hline
    WikiSmall &BLEU&FKGL&SARI& Simplicity \\\hline
    PBMT-R & 81.11 & 8.33&38.56&-0.55\\ \hline
    Hybrid & 48.97 & \textbf{4.56} & 31.40 & +0.36 \\ \hline
    SBMT-SARI  & 73.08 & 7.29& \textbf{39.96}& +0.03\\ \hline

    NMT  & 85.05 & 8.26 &36.14 & +0.30\\ \hline
    Dress & 77.18 & 6.58 & 37.08 & +0.25 \\ \hline
    NMT+synthetic & \textbf{87.16} & 6.56 &37.21 & \textbf{+0.42}\\
    \hline\end{tabular}
\end{table}

\begin{table}
  \centering
  \caption{The evaluation results on WikiSmall. The best results are highlighted in bold on each metric.}
  \begin{tabular}{|c|c|c|c|c|} \hline
    WikiSmall &BLEU&FKGL&SARI& Simplicity \\\hline
    PBMT-R & 46.31 &11.42 &15.97 & -0.36 \\ \hline
    Hybrid & 53.94 & 9.20 & 30.46 & +0.23\\ \hline

    SBMT-SARI  & 48.76 & 11.18& \textbf{32.75}& +0.02\\ \hline

    NMT  &  46.89 & 12.01 & 26.42 & +0.05\\ \hline
    Dress & 34.53 & \textbf{7.48}& 27.48 & +0.13 \\ \hline
    NMT+synthetic & \textbf{55.26} & 10.75 &  26.49& \textbf{+0.28} \\
    \hline\end{tabular}
\end{table}

\subsection{Conclusion}

In this paper, we propose one simple method to use simplified corpora during training of NMT systems, with no changes to the network architecture. In the experiments on two datasets, we achieve substantial gains in all tasks, and new SOTA results, via back-translation of simplified sentences into the ordinary sentences, and treating this synthetic data as additional training data. Because we do not change the neural network architecture to integrate simplified corpora, our method can be easily applied to
other Neural Text Simplification (NTS) systems. We expect that the effectiveness of our method not only varies with the quality of the NTS system used for back-translation, but also depends on the amount of available parallel and simplified corpora. In the paper, we have only utilized data from Wikipedia for simplified sentences. In the future, many other text sources are available and the impact of not only size, but also of domain should be investigated.

\bibliographystyle{acl}
\bibliography{coling2018}

\end{document}